\documentclass[10pt,a4paper]{article}

\usepackage{lrec-coling2024}

\usepackage{url}
\usepackage{comment}
\usepackage{amsmath,amssymb}
\usepackage{latexsym}
\usepackage{tabularx,booktabs}
\usepackage{hyperref}
\usepackage{cgloss4e}
\usepackage{multirow}
\usepackage{linguex}
\usepackage{graphicx}
\usepackage{xcolor}
\usepackage{pifont}
\usepackage{makecell}
\usepackage{CJKutf8}

\newcolumntype{Y}{>{\centering\arraybackslash}X}

\newcommand{\ours}{J-CRe3}
\newcommand{\cmark}{\ding{51}}
\newcommand{\xmark}{\ding{55}}
\newcommand{\header}[1]{\multicolumn{1}{c}{\textbf{#1}}}
\newcommand{\mc}[3]{\multicolumn{#1}{#2}{#3}}
\newcommand*{\Ja}[1]{\begin{CJK}{UTF8}{ipxm}#1\end{CJK}}

\definecolor{myorange}{rgb}{0.99,0.59,0.42}
\definecolor{mygreen}{rgb}{0.34,0.77,0.33}
\definecolor{myblue}{rgb}{0.38,0.66,0.86}

\title{
  J-CRe3: A Japanese Conversation Dataset \\
  for Real-world Reference Resolution
}

\name{
  \fontsize{11pt}{11pt}\selectfont
  Nobuhiro Ueda${}^{1,2}$,
  Hideko Habe${}^{2}$,
  Yoko Matsui${}^{2}$,
  Akishige Yuguchi${}^{3,2}$, \\
  \fontsize{11pt}{11pt}\selectfont
  \textbf{
    Seiya Kawano${}^{2,4}$,
    Yasutomo Kawanishi${}^{2,4}$,
    Sadao Kurohashi${}^{1,2,5}$,
    Koichiro Yoshino${}^{2,4}$
  }
}

\address{
  ${}^{1}$Kyoto University, Kyoto, Japan,
  ${}^{2}$Guardian Robot Project, R-IH, RIKEN, Kyoto, Japan, \\
  ${}^{3}$Tokyo University of Science, Tokyo, Japan,
  ${}^{4}$Nara Institute of Science and Technology, Nara, Japan, \\
  ${}^{5}$National Institute of Informatics, Tokyo, Japan \\
  {\{ueda,kuro\}@nlp.ist.i.kyoto-u.ac.jp} \hspace{3mm}
  {akishige.yuguchi@rs.tus.ac.jp} \\
  {\{hideko.habe,yoko.matsui,seiya.kawano,yasutomo.kawanishi,koichiro.yoshino\}@riken.jp}
}

\newcounter{exa}
\newcommand{\itemex}{\refstepcounter{exa}\item[\textmd{\textrm{(\theexa)}}]}
\def\exa#1{
  \begin{description}
    \itemex #1
  \end{description}
}

\abstract{
  Understanding expressions that refer to the physical world is crucial for such human-assisting systems in the real world, as robots that must perform actions that are expected by users.
  In real-world reference resolution, a system must ground the verbal information that appears in user interactions to the visual information observed in egocentric views.
  To this end, we propose a multimodal reference resolution task and construct a Japanese Conversation dataset for Real-world Reference Resolution (J-CRe3).
  Our dataset contains egocentric video and dialogue audio of real-world conversations between two people acting as a master and an assistant robot at home.
  The dataset is annotated with crossmodal tags between phrases in the utterances and the object bounding boxes in the video frames.
  These tags include indirect reference relations, such as predicate-argument structures and bridging references as well as direct reference relations.
  We also constructed an experimental model and clarified the challenges in multimodal reference resolution tasks.
  \\
  \newline 
}

\begin{document}

\maketitleabstract

\section{Introduction}

\begin{figure*}[t]
  \centering
  \includegraphics[width=0.95\textwidth]{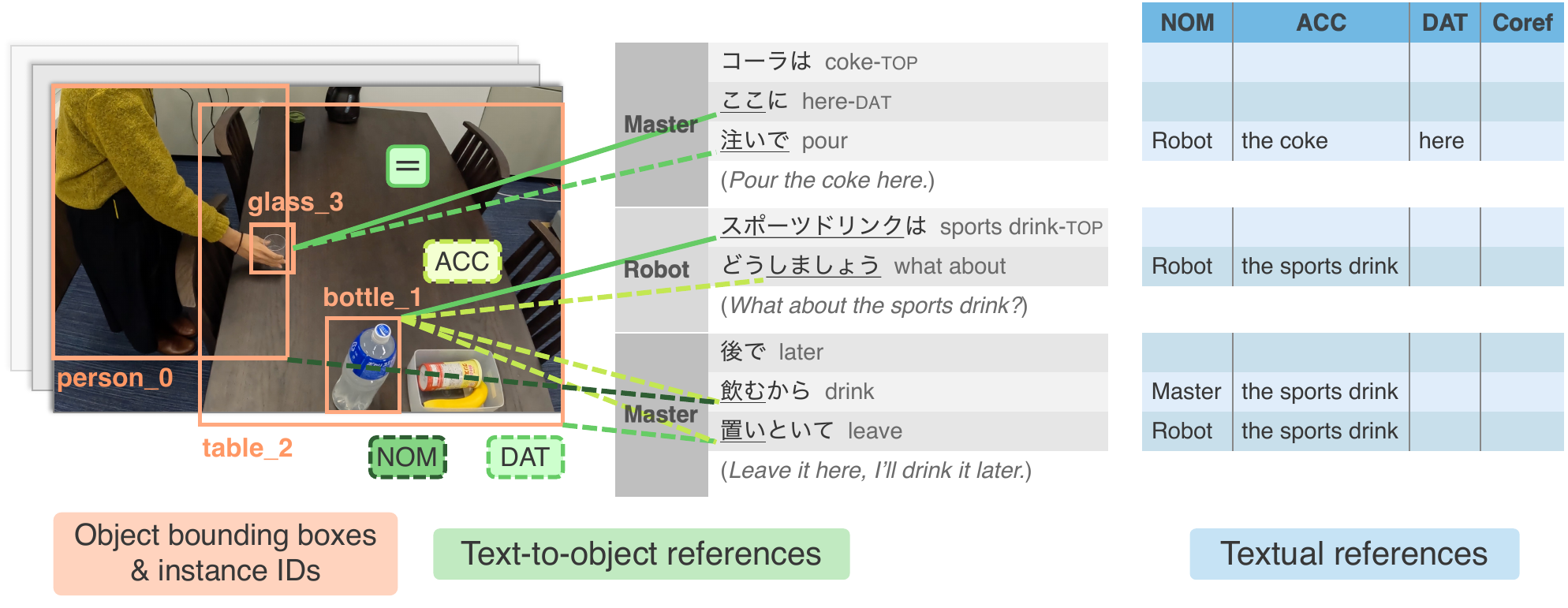}
  \caption{
    Example of \ours.
    It has object bounding boxes (\textcolor{myorange}{the orange rectangles}), textual reference relations (\textcolor{myblue}{the blue table}), and text-to-object reference relations (\textcolor{mygreen}{the green lines}).
    An object bounding box has a class name and an instance ID.
    Textual and text-to-object reference relations have 10--20 types of relations, including direct reference relations (=) and indirect reference relations corresponding to nominative (NOM), accusative (ACC), and dative (DAT) cases.
    For example, \textit{sports drink} has a direct text-to-object reference relation (=) with the object bounding box, ``bottle\_1.''
    Note that a particular case TOP shown in the example dialogue indicates an attached noun phrase is the sentence's topic.
  }
  \label{fig:overview}
\end{figure*}

% ロボットのように実世界で人間を補助するシステムにとって，身体世界を含めた照応表現の理解は重要な課題である．
% % 例えば人間と対話を通して協働するロボットにおいては，人間の発話内のフレーズを実世界のエンティティへ接地することが必須となる．
% 例えば，図~\ref{fig:overview}中の ``Pour the coke here'' というユーザ発話を例にとって考える．
% このとき， ``pour''というpredicateの格要素（nominative: robot, accusative: coke, dative: here=glass）を，実世界の実体に紐づけて正しく理解することが必要である．
% このとき， ``the coke,'' ``here'' などエンティティや照応表現，ひいては ``pour'' の動作主を実世界における実体と正しく結び付けなければロボットの動作に反映することができない．
%Human-assisting systems such as robots must understand relationships between language interaction and things in the physical world, as real-world reference resolution task.
Human-assisting systems such as robots will be active in our living spaces in the near future.
Such systems must understand the intention of users in the real world by grounding the referential expressions in language to real-world objects for cooperative action generation.
Take the utterance, \textit{Pour the coke here} as an example (Figure~\ref{fig:overview}).
For a robot to generate an appropriate action, the following arguments of predicate \textit{pour} must be recognized: nominative: \textit{robot}; accusative: \textit{the coke}; and dative: \textit{here}.
Furthermore, it is crucial to ground entities (\textit{the coke}), referential expressions (\textit{here}), and even the agent of \textit{pour} to their corresponding real-world entities.

% 発話中の表現と視覚情報を紐付けた対話データセットとして SIMMC 2.1~\cite{kottur-etal-2021-simmc,kottur-moon-2023-overview}がある．
% SIMMC 2.1 は1,566枚の CG 画像と11,244の対話テキストに対してテキスト中の名詞句と参照先の画像中の物体を囲む矩形が紐付けられたデータセットである．
% 規模は大きいものの， SIMMC 2.1 は話者の写っていない静止画に基づいており，物体の移動や操作が視覚的に提示されていない．
% これはSIMMCがバーチャル空間におけるインタラクションを指向しているためであるが，実世界におけるインタラクションにおいては話者や動作主と物体の関係は重要である．
% また， SIMMC 2.1 の参照関係はテキストとして表出している名詞句にのみ付与されている．
% 一方，日本語ではゼロ照応とよばれる参照元の表現が省略される現象が頻出する．
% 例えば「こっちに持ってきて (Can you bring (it) over here?)」という発話では，持ってくる対象が省略されている．
% こうしたゼロ照応も実世界における対話では頻繁に生じるため，考慮する必要がある．

Such reference resolution tasks with an egocentric view have been considered by existing works.
\citet{shirai-etal-2022-visual} proposed a dataset to understand the cooking procedures by bridging the recipe texts and cooking videos.
In an interactive scenario, SIMMC 2.1~\citep{kottur-etal-2021-simmc,kottur-moon-2023-overview} is a multimodal dialogue dataset that links referential expressions in dialogues with visual information in the virtual world.

In SIMMC 2.1, agents do not appear, and the movements or manipulation of objects are implemented as conceptual actions.
This is because the dataset is oriented toward interaction in virtual space, although the physical relations between agent and objects remain important in a real-world interactions.

Another issue is that SIMMC 2.1 only focuses on limited reference relations; it has reference relation annotations only for phrases that appear in texts.
However, referential phrases are frequently omitted in Japanese, called zero reference or zero anaphora~\citep{sasano-etal-2008-fully}.
For example, in the utterance, ``\Ja{こっちに持ってきて}'' (Can you bring (it) over here?\footnote{The indicator ``(it)'' is originally omitted in Japanese; it has been added to the English translation.}), the object to be brought is omitted.

% 本研究では，実世界での物体操作を伴う対話においてゼロ照応も総合的に扱うマルチモーダル参照解析を提案し，そのためのデータセットJ-CRe3: A Japanese Conversation Dataset for Real-world Reference Resolutionを構築する．
% データセット構築のため，まずクラウドソーシングで家庭内における主人とロボットの対話シナリオテキストを収集する．
% 続いて，収集したシナリオに沿って主人役とロボット役の演者が実世界で対話音声と対話動画を収録する．
% 最後に，対話音声の書き起こしと対話動画の1秒ごとのフレームに物体矩形および参照関係を付与する．
We propose a multimodal reference resolution task that comprehensively handles zero references in real-world conversations involving object manipulation tasks.
We constructed a dataset, J-CRe3: A Japanese Conversation dataset for Real-world Reference Resolution, which contains egocentric video and dialogue audio of real-world conversations between two people.
The conversations involve a robot that is helping its master with daily mundane tasks, including many object manipulations.
In addition, because the conversations are in Japanese, they naturally contain numerous zero references.

% 構築したコーパスとアノテーションの具体的例を図~\ref{fig:overview}に示す．
% 物体矩形にはクラス名およびインスタンスIDが付与される．
% 参照関係はテキスト間照応関係とテキスト・物体間参照関係の2種類がある．
% テキスト間照応関係は対話テキスト中のフレーズ間の述語項構造や共参照関係を含む．
% テキスト・物体間参照関係は図の「スポーツドリンク」のように，対話テキスト中の名詞句とそれが参照している物体矩形との関係である．
% 加えて本研究ではゼロ照応も扱うため，テキスト中に言及がない場合でもテキスト中の述語からそのガ格やヲ格に対応する物体を紐付ける．
% 我々のデータセットは1,684件の発話を含む93件の対話動画から構成される．
% % 我々のデータセットは53件の対話動画および1,159件の発話を含む対話書き起こしテキストから構成される．
% % 対話数： 53 + 37 + 1 + 1 + 1, 発話数：1159 + 495 + 10 + 10 + 10
Figure~\ref{fig:overview} shows an example of our dataset.
Each bounding box has a class name and an instance ID.
Our dataset has two types of reference relations: textual and text-to-object reference relations.
Textual reference relations include predicate-argument structures, bridging reference relations, and coreference relations.
The text-to-object reference relation is a connection between a noun phrase and the bounding box to which it refers, as in the case of \textit{here} and \textit{sports drink} in the example.
In addition, as this study handles zero references, we associate a predicate with the bounding boxes corresponding to its arguments, even when those arguments are omitted.
In Figure 1, the predicate \textit{leave} and its dative argument exemplify zero reference.
Although the dative argument of \textit{leave} does not appear in the text,\footnote{\textit{here} is omitted in the original Japanese text.}
\textit{leave} and the object ``table\_2'' are connected by the dative case relation, which the omitted argument would have with the predicate.

Our dataset consists of 93 videos and dialogue audio containing 2,131 utterances.
The number of dialogues in our dataset is relatively small compared to the other related datasets (Table~\ref{tab:dataset-comparison}).
However, our dataset has a fairly large number of unique images, in which all the objects referred to in the whole dialogue are densely annotated.

We also constructed an experimental model for our proposed multimodal reference resolution task to clarify the difficulty of the proposed task and the dataset.
The proposed task can be divided into three widely studied tasks: textual reference resolution, object detection, and text-to-object reference resolution.
Our experimental results showed that the accuracy of textual reference resolution was roughly the same as existing monologue datasets (F-scores of around 0.8).
However, the text-to-object reference resolution task was demonstrated to be challenging (recall of around 0.4) and to have much room for improvement.
Our dataset, including videos, audio, transcriptions, and annotations, is publicly available.\footnote{\url{https://github.com/riken-grp/J-CRe3}}
The source code and the weights of the resolution models used in this study are also publicly available.\footnote{\url{https://github.com/riken-grp/multimodal-reference}}

\begin{table*}
  \centering
  \small
  \scalebox{0.9}{
    \begin{tabular}{l|rcrcc}
      \toprule
      \mc{1}{c|}{\textbf{Dataset}}                            & \mc{1}{c}{\textbf{\makecell{\# Annotated                                                     \\ images}}} & \mc{1}{c}{\textbf{Text type}} & \mc{1}{c}{\textbf{\# Dialogues}} & \mc{1}{c}{\textbf{Video}} & \mc{1}{c}{\textbf{\makecell{Zero \\ reference}}} \\
      \midrule
      % ELDERLY-AT-HOME &
      RefCOCO~\citep{yu-2016-refcoco}                         & 20k                                      & Referring expression   & -      & \xmark & \xmark \\
      RefCOCO+~\citep{yu-2016-refcoco}                        & 142k                                     & Referring expression   & -      & \xmark & \xmark \\
      RefCOCOg~\citep{mao-2016-refcocog}                      & 26k                                      & Referring expression   & -      & \xmark & \xmark \\
      VisualGenome~\citep{krishna-2016-visualgenome}          & 108k                                     & Caption                & -      & \xmark & \xmark \\
      Flickr30k Entities~\citep{flickrentitiesijcv}           & 30k                                      & Caption                & -      & \xmark & \xmark \\
      VisCoref~\citep{yu-etal-2019-see}                       & 5k                                       & Dialogue               & 5,000  & \xmark & \xmark \\  % 5,000
      Visual Recipe Flow~\citep{shirai-etal-2022-visual}      & 6k                                       & Cooking recipe         & -      & \xmark & \xmark \\  % 5,659
      BioVL2~\citep{nishimura-2021-iccvw,nishimura-2022-jnlp} & 3k                                       & Experimental procedure & -      & \cmark & \xmark \\  % 2,521
      % VFD~\cite{kamezawa-etal-2020-visually}                 & 34,775  & 対話        & $\sim$389k & NO   & NO     \\
      EPIC-KITCHENS~\citep{Damen2022RESCALING}                & 277k                                     & Narration              & -      & \cmark & \xmark \\
      RefEgo~\citep{Kurita_2023_ICCV}                         & 226k                                     & Referring expression   & -      & \cmark & \xmark \\  % 226,319
      SIMMC 2.1~\citep{kottur-moon-2023-overview}             & 2k                                       & Dialogue               & 11,244 & \xmark & \xmark \\  % 1,566
      \midrule
      \textbf{\ours} (ours)                                   & 11k                                      & Dialogue               & 93     & \cmark & \cmark \\  % 11,062
      \bottomrule
    \end{tabular}
  }
  \caption{
    % テキスト中のフレーズと物体間の関係が付与された画像あるいは1人称視点動画データセットの比較
    Comparison of image or egocentric video datasets with relations between phrases and objects.
  }
  \label{tab:dataset-comparison}
\end{table*}

\section{Multimodal Reference Resolution}

% 本研究では，実世界での対話的協働を想定したマルチモーダル参照解析タスクを提案する．
% 本タスクでは，動画あるいは画像と，対応するテキストが入力として与えられる．
% そのとき，テキスト中の名詞や述語における参照関係を，参照先として動画や画像中の物体も含めて解析する．
% 本タスクは，テキスト間照応解析，物体検出，テキスト・物体間参照解析の3つのサブタスクから構成される．
We propose a multimodal reference resolution task for real-world interactive systems that collaborate with humans.
Given an egocentric image and the corresponding text as input, this task seeks the referents of nouns and predicates from objects in the image as well as phrases in the text.
This task consists of three subtasks: textual reference resolution (Section \ref{sec:textual-reference-resolution}), object detection (Section \ref{sec:object-detection}), and text-to-object reference resolution (Section \ref{sec:text-to-object-reference-resolution}).

\subsection{Textual Reference Resolution}\label{sec:textual-reference-resolution}
% \textbf{テキスト間照応解析}は，テキスト中の単語や句の間に存在する照応・共参照関係を解析するタスクである．
% 本研究では先行研究~\cite{ueda-2020,umakoshi-etal-2021-japanese-zero,Ueda-2023-kwja}にならい，述語項構造，共参照，橋渡し照応関係を解析する．
% 述語項構造は述語を中心とし，その述語の「誰が」や「何を」に相当する項からなる関係である．
% 図~\ref{fig:overview}では「leave」と「drink」とについて述語項構造が示されている．
% 共参照関係は実世界において同一の実体を指し示す名詞間の関係である．
% 橋渡し照応関係はある名詞（照応詞）と，その必須的な意味を補完する異なる名詞（先行詞）との関係である．
Textual reference resolution recognizes semantic relations among phrases in a text.
Following previous studies~\citep{ueda-2020,umakoshi-etal-2021-japanese-zero,Ueda-2023-kwja}, we focus on predicate-argument structure (PAS), coreference, and bridging reference.

PAS is a set of relations between a predicate and its arguments, which correspond to \textit{who} did/does \textit{what} to \textit{whom} for the predicate.
Figure~\ref{fig:overview} (right) shows that the predicate \textit{leave} has two arguments, \textit{Robot} and \textit{the sports drink}.
Note that here, the accusative case \textit{it} for the predicate \textit{leave} is omitted, and instead, \textit{the sports drink} is marked as the accusative case.
Because the omitted argument \textit{it} refers to \textit{the sports drink}, this is an example of zero reference, and thus PAS is used to annotate zero references~\citep{kwdlc-paclic-2012}.

Coreference is a phenomenon where two (or more) noun phrases refer to the same entity in the real world.
A bridging reference relation is an indirect relation between two noun phrases where one noun phrase (anaphor) refers to the other (antecedent) and the latter complements the essential meaning of the former.
These semantic relations are all crucial for dialogue understanding by interactive robots operating in the real world.

\subsection{Object Detection}\label{sec:object-detection}
% \textbf{物体検出}は，画像中から参照されている物体が存在する領域を特定するタスクである．
% 図~\ref{fig:overview}においては図中の物体矩形を推定することに対応する．
% 入力が動画の場合は動画中のそれぞれのフレームに対して同様の処理を行う．
Object detection identifies and locates objects within an image.
The output of this task is object bounding boxes, as shown in Figure~\ref{fig:overview} (left).
The detected bounding boxes are fed to the text-to-object reference resolution task.

\subsection{Text-to-object Reference Resolution}\label{sec:text-to-object-reference-resolution}
% \textbf{テキスト・物体間参照解析}は，テキスト間照応解析における照応・共参照の対象を物体検出によって特定された物体領域から選択するタスクである．
% 図~\ref{fig:overview}においては単語と物体矩形を結ぶエッジを推定することに対応する．
% 表現が直接参照している対象を画像中から検出するタスクはphrase grounding~\cite{kamath2021mdetr,gupta2020contrastive}や referring expression comprehension~\cite{qiao2020referring}として知られる．
% しかし，視覚情報が共有されていることからフレーズが頻繁に省略される実世界の対話においては，direct reference の解決だけでは発話理解には至らない．
% したがって，本研究では述語項構造や橋渡し照応など，ゼロ照応を含む間接的な関係も扱う．
Text-to-object reference resolution identifies the referents of nouns and predicates, similar to textual reference resolution.
Yet, the referents are selected from the object detection's output.
In Figure~\ref{fig:overview}, this task predicts the edges between the words and the object bounding boxes.
The task of detecting an object directly referenced by a phrase is known as phrase grounding~\citep{kamath2021mdetr,gupta2020contrastive} or referring expression comprehension~\citep{qiao2020referring}.
However, in real-world conversations, where the interlocutors share the visual information and phrases are frequently omitted, resolving direct references is insufficient for adequate comprehension of their utterances.
Thus, we also consider indirect relations, including PAS and bridging references, which involve zero references.

\section{\ours ~Dataset}

% 本研究では，マルチモーダル参照解析のための\ours データセットを構築する．
% 本データセットは，実世界における主人役とロボット役2者の対話シーンにおいて動画，音声を収録し，音声書き起こし，照応・参照関係アノテーションを付与したデータセットである．
% 対話内容は，人間とお手伝いロボットの対話を想定する．
% 対話場面は，家庭内のリビング，ダイニング，キッチンを模した3種類である．
% 本データセットには93対話が含まれるが，テキスト間照応解析アノテーション済みのものは53対話，フルアノテーション済みのものは30対話である．
% 本節では\ours データセットの構築方法について順に述べる．
We constructed the \ours~dataset for multimodal reference resolution.
It consists of video and audio recordings of real-world conversation scenes between two people, audio transcriptions, and annotations of various reference relations.
Through the lens of applications to human-assisting systems, we assumed that the two interlocutors are a master and an assistant robot and prepared three dialogue locations: a living room, a dining room, and a kitchen.

In this section, we describe the construction procedure and statistics of \ours.
First, we collected dialogue scenarios through crowdsourcing (Section~\ref{sec:scenario-collection}).
Then, we recruited actors for the master and robot roles and recorded the egocentric videos and the dialogue audio of their conversations following the collected scenarios (Section~\ref{sec:conversation-recording}).
Finally, we labeled the bounding boxes and the reference relations to the audio transcriptions and the video frames per second (Section~\ref{sec:annotation}).
The statistics of the dataset are described in Section~\ref{sec:statistics}.

\subsection{Dialogue Scenario Collection}\label{sec:scenario-collection}

% 多様かつ現実的な対話シナリオを得るため，クラウドソーシングを利用してシナリオを収集した．
% クラウドソーシングタスクではワーカーに対話収録に使用する部屋の状況と使用可能な物体の写真を提示した．
% その上で，人間とロボットの発話およびその際の動作や場面状況を記述してもらった．
% 発話数は長すぎず，かつ対話が十分な文脈を持つよう10–16発話に制限した．
% 収集したシナリオを実行可能性，十分な頻度の参照表現，十分な粒度の場面状況説明，の3つの観点からフィルタリングし，残ったシナリオを自然な対話になるよう修正した．
% 付録~\ref{sec:scenario-example}に修正後のシナリオの例を示す．
We collected a variety of realistic dialogue scenarios through crowdsourcing.
In the crowdsourcing task, the workers were shown pictures of the room and objects to be used in the conversation recording.\footnote{The crowdsourcing interface is shown in Appendix~\ref{sec:crowdsourcing-interface}.}
The workers then wrote dialogue texts along with the interlocutors' actions and surrounding situations.
101 workers participated in our task, and 180 scenarios were collected.
The number of utterances per scenario was limited to 10--16 to ensure that the dialogues were not too long and had sufficient context.
We manually filtered out the collected scenarios that lacked feasibility, a sufficient number of referential expressions, and sufficient descriptive granularity for the situation and manually modified the remaining scenarios for more naturalness.
Appendix~\ref{sec:scenario-example} shows an example of a modified scenario.

\subsection{Conversation Recording}\label{sec:conversation-recording}
% 収集したシナリオに基づいて実際に対話を行い，データを収録した．
% シナリオは人対ロボットを想定しているが，今回はロボット役も人間の演者に依頼した．
% 演者にはできる限りシナリオを暗記してもらい，対話中の振る舞いが自然になるようにした．
% なお，データセットにおいては実際に行われた発話に対応するよう，アノテーションの際にシナリオを元に台詞を修正する．
We recruited five actors and paired two of them to perform the master and robot roles and recorded their in-person conversations following the modified scenarios.
% We asked them to memorize the scenario for recording a natural conversation.
% 収録はリビングとダイニングとキッチンを模した設備が備え付けられた実験室で行った．
% 2人の演者にはそれぞれピンマイクを付けてもらい，発話を録音した．
% ロボット役の演者には頭部にカメラ\footnote{GoPro HERO10 Black}を付けてもらい，対話中の1人称視点動画を撮影した．
% さらに，実験室に定点カメラを4箇所設置し，部屋全体の様子を撮影した．
The recording was conducted in a laboratory furnished to resemble a living room, a dining room, and a kitchen.\footnote{The furnished recording area is a room of approximately 6.5m x 6.8m in which the three locations are set up without walls.}
Both actors were equipped with close-talking microphones to record their speeches.
The actor playing the robot had a head-mounted RGB camera\footnote{
  We used GoPro HERO10 Black.
  It has a wide-angle lens and can capture the whole room from the corner.
}
to capture an egocentric video during the conversation.
We installed four fixed RGB cameras in the corner of the ceiling in the laboratory to record third-person videos, \textit{i.e.,} the entire room's overview.

\subsection{Annotation}\label{sec:annotation}
% 収録した対話音声と1人称動画に対してマルチモーダル参照解析のためのアノテーションを行った．
% 3人称視点動画にはアノテーション対象となる物体が十分な大きさで写っていないことが多かったため，タグ付けの際に参照する程度の利用にとどめた．
% まず対話音声は発話単位でテキストに書き起こし，1人称視点動画は1秒ごとにフレームを抽出し画像系列に変換した．\footnote{
%   An \textit{utterance} is almost always a series of sentences uttered by a speaker before the turn transition.
%   However, in the case of a utterance with long pause, the utterance may split into multiple utterances.
% }
% 書き起こしの際には，動画との対応が取れるよう発話の開始時間と終了時間を記録した．
We annotated the collected conversational audio and egocentric videos for multimodal reference resolution.
The third-person videos were used only as a reference of the objects that are occluded or out of view in the egocentric videos.
%because the objects recorded by these cameras were not sufficiently large.

We transcribed every utterance\footnote{
  An \textit{utterance} is almost always a series of sentences made by a speaker before a turn transition.
  However, following~\citet{yoshino-etal-2018-japanese}, when an utterance's pause exceeds 500 msec and the utterance's semantic content is complete at that point, an utterance is delimited.
}
from the conversational audio and converted the egocentric videos into image sequences by extracting the frames every second.
We annotated each utterance with timestamps at its beginning and ending points to ensure proper alignment with the videos.

% 以下では，テキスト間照応解析，物体検出，テキスト・物体間参照解析に対応するアノテーションをそれぞれテキスト間照応アノテーション，物体領域アノテーション,テキスト・物体間参照アノテーションとよぶ．
The following sections describe the annotation for the textual reference resolution, the object detection, and the text-to-object reference resolution.

\subsubsection{Textual Reference Annotation}
% 書き起こされた対話テキストに対して述語項構造・共参照・橋渡し照応関係を付与した．
% アノテーションの基準は京都大学ウェブ文書リードコーパス~\citep{hangyo-etal-2012-building}に準拠した．
% 本データセットに含まれるテキストは対話形式の話し言葉である一方，京都大学ウェブ文書リードコーパスにはモノローグ形式の書き言葉が含まれる．
% そのため，アノテーションの際には追加の基準を設けた．
% アノテーションにおいては事前に\citet{ueda-2020}の単語選択モデルでシルバーアノテーションを付与した．
We annotated the transcribed conversational text with predicate-argument structures, coreference relations, and bridging reference relations.
We followed the annotation guidelines of an existing textual reference corpus~\citep{kwdlc-paclic-2012}.
Although the existing corpus contains monologue-style written text, our dataset contains dialogue-style spoken text.
Therefore, we defined additional guidelines for colloquial expressions, including casual replies.\footnote{\url{https://github.com/riken-grp/J-CRe3/blob/main/docs/annotation_guideline.pdf}}

In Figure~\ref{fig:overview}, coreference relation would be annotated between \textit{the sports drink} and \textit{it}, if \textit{it} were mentioned in the dialogue.
For bridging reference, if Robot said, ``Drink it early because the expiration date is approaching,'' the relation would be annotated between \textit{the sports drink} and \textit{the expiration date}.

\subsubsection{Bounding Box Annotation}
% 1人称視点動画から抽出された各画像に対し，物体矩形を付与した．
% また，それぞれの物体矩形に対して物体のクラス名およびインスタンスIDを付与した．
% クラス名は物体認識タスクにおいて広く利用される LVIS データセット~\citep{gupta2019lvis}において定義されている 1,203 クラスの集合から選択する．
% インスタンスIDはそれぞれの物体を一意に識別するためのIDである．
% インスタンスIDは動画中で一貫して付与される必要がある．
% すなわち，同じ物体が異なる画像フレームに出現した場合は同じインスタンスIDを付与する．
We annotated each image extracted from the egocentric videos with object bounding boxes.
We also labeled an object class name and an instance ID for each bounding box.
The class name was selected from a set of 1,203 classes defined in the LVIS dataset ~\citep{gupta2019lvis}, which is widely used in object detection tasks.
An instance ID is an identifier that uniquely distinguishes each object and must be assigned consistently throughout the video.
% In other words, the same instance ID must be assigned when the same object appears in different frames.

% 作業においては，一般物体認識器 Detic~\citep{zhou2022detecting}の学習済みモデル\footnote{\url{https://github.com/facebookresearch/Detic/blob/main/docs/MODEL_ZOO.md}}と複数物体追跡器SORT~\citep{Bewley2016_sort}を使用してシルバーアノテーションを事前付与し，手作業で修正した．
We used a general object detector called Detic~\citep{zhou2022detecting} and a multi-object tracker called StrongSORT~\citep{strongsort} to automatically annotate the bounding boxes, which were then manually corrected.

% \paragraph{インスタンスIDのアノテーション}

% それぞれの物体矩形に対して物体を一意に識別するためにインスタンスIDを付与する．
% インスタンスIDは物体ごとに固有でなければならない．
% 例えば，同じ色の皿が2枚ある場合には，それぞれに異なるインスタンスIDを付与する．

% インスタンスIDは動画中で一貫して付与される必要がある．
% すなわち，同じ物体が異なる画像フレームに出現した場合は同じインスタンスIDを付与する．
% ただし，似た物体が複数同時に出現するなどの理由で，動画から特定の物体の追跡が困難な場合には，可能性のあるインスタンスIDから任意に選択して付与する．

\subsubsection{Text-To-Object Reference Annotation}
% 我々はテキスト中の (1) 名詞句および (2) 述語と，画像中の物体矩形のすべての組み合わせについて参照関係を付与した．
% (1) 名詞句については，直接参照している物体および橋渡し照応関係にある物体に矩形を付与した．
% (2) 述語についてはその項に対応する物体矩形を格ごとに付与した．
% この基準は，テキスト間照応アノテーションにおける照応先を物体矩形に拡張したものに相当する．
We assigned reference relations to every combination of phrases in the text and bounding boxes in the image.
The phrases include noun phrases and predicates.
For noun phrases, we assigned bounding boxes to objects directly referenced or with bridging reference relations.
For predicates, we assigned bounding boxes to objects corresponding to the predicates' arguments.
Text-to-object reference annotation is similar to the textual reference annotation where the reference target is extended to object bounding boxes.

% テキスト・物体間参照アノテーションは付与対象の関係が非常に多くなる．
% しかし，付与済みのインスタンスIDを利用することで大部分のアノテーションを省くことができる．
% 例えば，ある動画フレーム中の「コップ」に対して参照関係を付与した場合を考える．
% このとき，別フレームに同じインスタンスIDを持つ「コップ」が出現したとしても自動的に参照関係を付与できる．
% また，付与済みのテキスト間照応・共参照関係も利用できる．
% 例えば，以下のようなテキスト間照応アノテーションが付与済みだった場合を考える．
Text-to-object reference annotation involves a significantly large number of relations to be annotated.
However, we can eliminate most of them by utilizing previously assigned instance IDs.
For example, given that a reference relation is assigned to a bounding box of a ``cup'' in a specific video frame, the reference relations of the same ``cup'' in other frames can be automatically assigned.
We can also utilize previously assigned textual reference relations.
For example, consider a case where the following text reference annotation has already been assigned to an utterance text:

% \exa{【Master】 そこにあるコップを、机に\underline{運んで} (ガ格:ロボット, ヲ格:コップ, ニ格:机)}
\exa{Can you \underline{put} that cup on the table? \\(NOM: robot, ACC: cup, LOC: the table)}

\noindent
% このとき，「コップ」とコップに対応する物体矩形に=格を付与すれば，「運んで」と「コップ」の物体矩形の関係は自動的にヲ格と推定できる．
% したがって，「運んで」のヲ格のタグ付けは不要である．
% 「机」についても同様である．
Once we assign a direct reference relation between \textit{cup} and a bounding box, the relation between the predicate \textit{put} and the bounding box can be automatically labeled as the accusative case.
Therefore, no accusative text-to-object annotation for \textit{put} is required.
The same applies to \textit{the table}.

% 20220304-56132132-0, 069.png
\begin{figure}[t]
  \centering
  \includegraphics[width=0.43\textwidth]{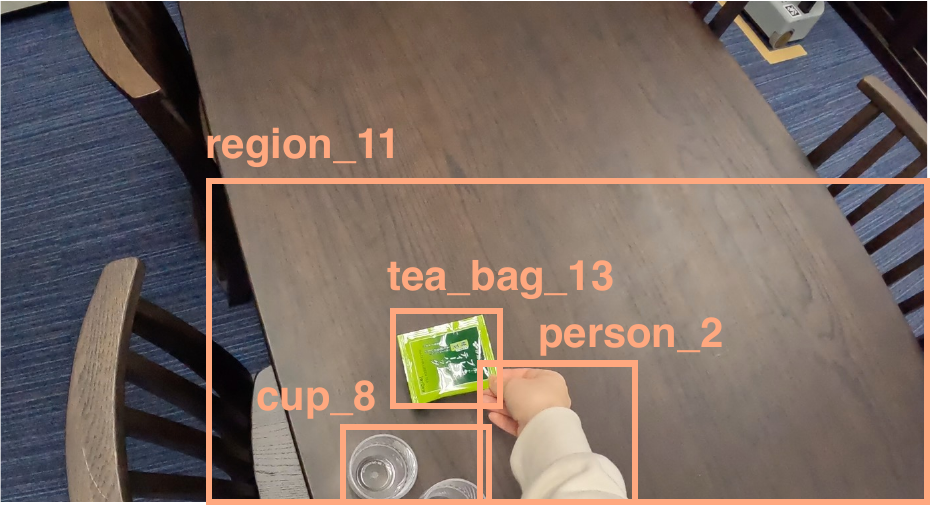}
  \caption{Example of regional bounding box annotation}
  \label{fig:region-1}
\end{figure}

% 実世界対話において特徴的な現象として「ここ」や「あそこ」など，場所を指す指示代名詞の使用が挙げられる．
% 指し示された場所の特定は，人間との協働において不可欠である．
% 本研究では，このような名詞を領域参照表現とよび，テキスト・物体間参照アノテーションにおいて対応する領域を付与する．
% この領域は物体領域とは異なり矩形が一意に定まらないため，その位置・大きさはアノテータの主観に依存する．
% そのため，特殊なクラス名である「region」を付与し，特殊な物体矩形として扱い，他の物体矩形とは区別する．
% 図~\ref{fig:region-1}に以下の発話についての領域矩形アノテーションの例を示す．
% 発話中の「そこ」という表現はいずれの物体にも対応せず，机の上の一部の領域を参照しているため，その領域を領域矩形としてタグ付けする．
% 同時に，この領域矩形と「そこ」に=格を付与する．
A characteristic phenomenon in real-world conversation is the use of demonstrative pronouns that refer to such locations, as \textit{here} and \textit{there}.
Identifying such indicated locations is essential for robots collaborating with humans.
In this study, we describe such phrases as regional referring expressions and assign corresponding regions to them in text-to-object reference annotation.
Unlike object bounding boxes, the bounding boxes of these regions are not uniquely determined, and their position and size depend on an annotator's subjective judgment.
For this reason, we assigned a special class name \textit{region} to these bounding boxes to distinguish them from other object bounding boxes.
Figure~\ref{fig:region-1} shows an example of regional bounding box annotation for the following utterance:

% \exa{【主人】 ついでに\underline{そこ}にあるコップとカップも片付けてちょうだい。 }
\exa{Can you put that cup away \underline{right there}?\\(=: region\_11)}

\noindent
As expression \textit{right there} does not refer to any object but rather to a part of the table, the region is tagged as a regional bounding box.
The regional bounding box and \textit{right there} are assigned a direct reference relation.

% python src/stat_text.py --knp-dir data/knp --id-dir data/id --table-fmt latex_booktabs
% python src/stat_image.py --annotation-dir data/image_text_annotation --knp-dir data/knp --id-dir data/id --table-fmt latex_booktabs
\begin{table}[t]
  \setlength{\tabcolsep}{5pt}
  \centering
  \small
  \scalebox{0.9}{
    \begin{tabular}{l|rrrr}
      \toprule
                           & \header{Train} & \header{Val.} & \header{Test} & \header{Total} \\
      \midrule
      \mc{4}{l}{\textbf{Recorded Data}}                                                      \\
      \midrule
      Dialogues            & 75             & 9             & 9             & 93             \\
      Utterances           & 1,746          & 155           & 230           & 2,131          \\
      Sentences            & 2,176          & 197           & 279           & 2,652          \\
      Morphemes            & 13,780         & 1,319         & 1,720         & 16,819         \\
      Total duration (sec) & 8,747          & 919           & 1,358         & 11,024         \\
      \midrule
      \mc{4}{l}{\textbf{Textual Reference Annotation}}                                       \\
      \midrule
      Predicates           & 2,780          & 264           & 342           & 3,386          \\
      Nominative args.     & 2,824          & 274           & 349           & 3,447          \\
      Accusative args.     & 1,138          & 94            & 135           & 1,367          \\
      Dative args.         & 1,399          & 133           & 146           & 1,678          \\
      Nominative-2 args.   & 527            & 63            & 62            & 652            \\
      Bridging anaphors    & 463            & 30            & 50            & 543            \\
      Coref. mentions      & 1,094          & 104           & 121           & 1,319          \\
      \midrule
      \mc{4}{l}{\textbf{Bounding Box Annotation}}                                            \\
      \midrule
      Frames               & 8,780          & 922           & 1,360         & 11,062         \\
      Bounding boxes       & 65,431         & 5,079         & 9,184         & 79,694         \\
      Object instances     & 1,435          & 119           & 187           & 1,741          \\
      Object classes       & 158            & 42            & 49            & -              \\
      \midrule
      \mc{4}{l}{\textbf{Text-To-Object Reference Annotation}}                                \\
      \midrule
      Direct reference     & 1,306          & 102           & 160           & 1,568          \\
      Nominative case      & 2,091          & 155           & 297           & 2,543          \\
      Accusative case      & 1,134          & 74            & 136           & 1,344          \\
      Dative case          & 1,228          & 96            & 148           & 1,472          \\
      Nominative-2 case    & 522            & 47            & 65            & 634            \\
      Bridging references  & 440            & 18            & 41            & 499            \\
      Zero references      & 6,016          & 453           & 708           & 7,177          \\
      \bottomrule
    \end{tabular}
  }
  \caption{Statistics of \ours.}
  \label{tab:statistics}
\end{table}

\subsection{Statistics}\label{sec:statistics}

Table~\ref{tab:statistics} shows the statistics of our dataset.
The number of object instances and classes indicates that our dataset contains bounding boxes of diverse objects.
The number of unique object classes in our entire dataset is 166.
The number of zero references is significantly larger than that of direct references, suggesting the importance of resolving zero references.

To quantify the textual diversity of our dataset, we calculated a dataset-level distinct-1 and distinct-2 scores~\citep{li-etal-2016-diversity}, where we counted unique n-grams over the entire dataset.
These scores of the dialogue texts in our dataset were 0.087 and 0.336, respectively, while those of the dialogue texts in SIMMC 2.1 were 0.054 and 0.285, indicating that our scenarios are more diverse than SIMMC 2.1.\footnote{
  We used the Japanese morphological analyzer, Juman++~\citep{tolmachev-2018} and the nltk toolkit~\citep{bird2009natural} to tokenize Japanese and English texts, respectively.
  For SIMMC 2.1 dataset, we randomly sampled utterances from the dev split to match the number of words in our dataset for fair comparison.
}

\section{Reference Resolution}
% 構築したデータセットを使用してモデルの学習及び評価を行う．
% マルチモーダル参照解析は，テキスト間照応解析，物体検出，テキスト・物体間参照解析解析の3つのサブタスクから構成される．
% 実験では，テキスト間照応解析，とその他2つのタスクをそれぞれ独立に学習・評価し，最後にそれらの結果を統合する．
We clarified the difficulty of the multimodal reference resolution task and the constructed dataset by training and evaluating an experimental model on it.
Multimodal reference resolution consists of three subtasks: textual reference resolution, object detection, and text-to-object reference resolution.
In our experiments, we independently trained and evaluated the textual reference resolution and the other two tasks and then combined the results.

\subsection{Textual Reference Resolution}

\begin{table*}[t]
  \centering
  \small
  \scalebox{0.9}{
    \begin{tabular}{ll|r@{\hskip 1mm}rr@{\hskip 1mm}rr@{\hskip 1mm}r|r@{\hskip 1mm}rr@{\hskip 1mm}rr@{\hskip 1mm}r}
      \toprule
      \mc{2}{c|}{\multirow{2}{*}{\textbf{Task}}}         & \mc{6}{c|}{\textbf{\ours}}         & \mc{6}{c}{\textbf{KWDLC}}                                                                                                                                                                                     \\
      \mc{2}{l|}{}                                       & \mc{2}{c}{\textbf{Endophora}}      & \mc{2}{c}{\textbf{Exophora}} & \mc{2}{c|}{\textbf{All}} & \mc{2}{c}{\textbf{Endophora}} & \mc{2}{c}{\textbf{Exophora}} & \mc{2}{c}{\textbf{All}}                                                              \\ \midrule
      PAS analysis                                       & \multicolumn{1}{|l|}{Nominative}   & 0.84                         & (210)                    & 0.86                          & (211)                        & 0.85                    & (422) & 0.87   & (3649) & 0.77  & (2269) & 0.83   & (5918) \\ \midrule
                                                         & \multicolumn{1}{|l|}{Accusative}   & 0.87                         & (164)                    & 0.06                          & (3)                          & 0.85                    & (167) & 0.86   & (2218) & 0.45  & (238)  & 0.82   & (2456) \\ \midrule
                                                         & \multicolumn{1}{|l|}{Dative}       & 0.89                         & (74)                     & 0.81                          & (118)                        & 0.84                    & (192) & 0.80   & (1239) & 0.64  & (582)  & 0.75   & (1822) \\ \midrule
                                                         & \multicolumn{1}{|l|}{Nominative-2} & 0.00                         & (6)                      & 0.89                          & (76)                         & 0.85                    & (82)  & 0.58   & (179)  & 0.57  & (104)  & 0.58   & (283)  \\ \midrule
      \multicolumn{2}{l|}{Bridging reference resolution} & 0.83                               & (56)                         & 0.41                     & (8)                           & 0.79                         & (64)                    & 0.69  & (1897) & 0.55   & (208) & 0.68   & (2106)          \\ \midrule
      \multicolumn{2}{l|}{Coreference resolution}        & 0.72                               & (71)                         & 0.60                     & (15)                          & 0.70                         & (86)                    & 0.84  & (1746) & 0.77   & (350) & 0.82   & (2097)          \\
      \bottomrule
    \end{tabular}
  }
  \caption{
    F-scores of textual reference resolution:
    Endophora is a reference to entities that appear in the text.
    Exophora is a reference to entities that do not.
    We fine-tuned our model with three different random seeds and report the mean performances.
    Numbers of gold references are shown in parentheses.
  }
  \label{tab:textual-reference-result}
\end{table*}

\subsubsection{Task Settings}
% テキスト間照応解析はフレーズ間の意味的関係を特定するタスクであり，述語項構造解析，橋渡し参照解析，共参照解析から構成される．
% 本実験では~\citet{ueda-2020,Ueda-2023-kwja}にならい，主人とロボット両者の発話書き起こしテキストに含まれるフレーズから以下の条件でフレーズを抽出し，学習および評価の対象とする\footnote{We used the Japanese morphological analyzer Juman++~\citep{tolmachev-2018} for part-of-speech tagging and the Japanese versatile dependency parser KNP~\cite{kurohashi-1994-knp} for phrase feature tagging.}．
Textual reference resolution consists of predicate-argument structure (PAS) analysis, bridging reference resolution, and coreference resolution.
Following \citet{ueda-2020,Ueda-2023-kwja}, we extracted predicates and eventive noun phrases for PAS analysis, non-eventive noun phrases for bridging reference resolution, and noun phrases for coreference resolution from the transcribed utterances.

Unlike bridging reference and coreference resolutions, PAS analysis classifies the relations between predicates and their arguments into a set of predefined labels, \textit{i.e.,} cases.
In this study, we focus on four cases: nominative, accusative, dative, and nominative-2.\footnote{
  Nominative-2 is used for a common Japanese construction in which a predicate has two nominative arguments.
}

\subsubsection{Method}

We employed a word selection model for textual reference resolution following previous works~\citep{ueda-2020,Ueda-2023-kwja}.
This model formulates all three tasks as a word selection task in which the model selects a word from a given text as the referent of a target word (e.g., predicate).
We added two layers of feed-forward neural networks for each task on top of a pre-trained encoder model\footnote{
  We used a DeBERTa V2 model pre-trained on Japanese corpora (\url{https://huggingface.co/ku-nlp/deberta-v2-large-japanese}).
} and fine-tuned the whole model.

For fine-tuning the model, we utilized the following textual reference resolution corpora: the Kyoto University Text Corpus~\citep{kyotocorpus-lrec-1998,kyotocorpus-lrec-2002},\footnote{\url{https://github.com/ku-nlp/KyotoCorpus}} the Kyoto University Web Document Leads Corpus (KWDLC, \citealp{kwdlc-paclic-2012}),\footnote{\url{https://github.com/ku-nlp/KWDLC}} the Annotated Fuman Kaitori Center Corpus,\footnote{\url{https://github.com/ku-nlp/AnnotatedFKCCorpus}} and the Wikipedia Annotated Corpus.\footnote{\url{https://github.com/ku-nlp/WikipediaAnnotatedCorpus}}
These corpora contain 9,207 documents, an amount that mitigates the data scarcity problem of our dataset.
We mixed the training split of our dataset with the existing corpora to train the model.

% これらデータセットと本データセットを混合して使用するため，我々は本データセットにおける話者のラベル（「主人」および「ロボット」）を「相対的」になるよう変換する．
% 既存のコーパスはテキスト中に表出するフレーズ間の関係だけでなく，テキストの「書き手」や「読み手」などテキスト中に表出しないエンティティとの関係（外界照応関係）も付与されている．
% これらの関係ラベルを活用するため，我々は本データセットに付与されている「ロボット」および「主人」という外界照応ラベルを，発話ごとに「話し手」あるいは「聞き手」に変換する．
% 実験では，この相対的なラベルを「書き手」や「読み手」ラベルと同一視することで既存のコーパスと本データセットを混合してモデルを学習する．
% なお，モデルは相対ラベルを予測するよう学習されるが，発話に付与された話者ラベルを使用することで，「主人」と「ロボット」の絶対ラベルに変換が可能である．\footnote{
%   We also examined the model using absolute labels of master and robots, but the relative labeling method achieved better scores in most cases.
% }
To combine these corpora with our dataset, we converted its speaker labels (``master'' and ``robot'') to be \textit{relative}.
Existing corpora provide textual reference relations not only between phrases but also between a phrase and an entity that does not appear in the text, such as ``writer'' and ``reader.''
Such references are called exophoras.
To utilize these labels, we transformed the exophora labels ``master'' and ``robot'' in our dataset into ``speaker'' or ``listener'' for each utterance.
We then treated these relative labels as equivalent to the ``writer'' and ``reader'' labels, allowing us to mix all the datasets.
Although the model is trained to predict the relative labels, they can be converted to absolute labels (``master'' and ``robot'') using the speaker labels assigned to each utterance.\footnote{
  We also examined a model using absolute labels of ``master'' and ``robot,'' although the relative labeling method achieved better scores in most cases.
}

\subsubsection{Results}

% 結果を表~\ref{tab:textual-reference-result}に示す．
% 日本語テキスト間照応解析のタスクとして標準的に用いられているKWDLCと比べ，いずれのタスクにおいても遜色ない結果が得られた．
% KWDLCとのスコアの違いで特筆すべき点として，ガ格，ニ格，ガ２格の外界照応の精度が高いことが挙げられる．
% ガ格やガ２格の項は主体が入ることが多く，\ours においては主体のほとんどが主人かロボットであるため解きやすいタスクになっていたと考えられる．
% ニ格も相手に何かをしてもらう場合にその相手が入ることが多く，同様の理由で精度が高くなったと考えられる．
% したがって，対話参与者が3者以上になった場合の精度についてはさらなる検証が必要である．
Table~\ref{tab:textual-reference-result} shows the experimental results.
In all the tasks, we achieved comparable results to those of KWDLC, a monologue web corpus widely used for Japanese textual reference resolution.
A notable difference between \ours~and KWDLC is the scores of exophora reference resolution in the nominative, dative, and nominative-2 cases.
Arguments of the nominative and nominative-2 cases generally include an agent; thus, they are easier to resolve when the agent is a master or a robot.
Also, arguments of the dative case often include a listener who is asked to do something, making the resolution easier due to the limited referents.
Therefore, further verification is needed to evaluate the performance when the number of interlocutors exceeds two.

\subsection{Object Detection and Text-To-Object Reference Resolution}
% テキスト・物体間参照解析のうち，直接の参照関係のみを扱ったタスクは phrase grounding とよばれる．
% 本実験では既存の phrase grounding モデルを使用した実験について述べる．
% なお，述語項構造や橋渡し照応に対応する間接的な参照関係は扱わない．
% また，領域参照表現の解析も行わない．
Text-to-object reference resolution can be divided into direct reference resolution (denoted as ``='' in Figure~\ref{fig:overview}) and indirect reference resolution (denoted as ``NOM'', ``ACC'', and ``DAT'').
The direct reference resolution is also called phrase grounding.
Although a phrase grounding model cannot resolve indirect references, including zero references, it is actively studied and many models and datasets have been proposed~\citep{kamath2021mdetr,gupta2020contrastive,flickrentitiesijcv,nakayama-tamura-ninomiya:2020:LREC}.
To investigate the extent to which these models can solve text-to-object reference resolution, even partially, this section discusses experiments with an existing phrase grounding model.
% We also investigate whether these datasets can be effectively used to solve the direct reference resolution task.

\subsubsection{Task Settings}
% phrase grounding は，テキストと画像が与えられたときテキスト中のフレーズに対応する画像中の物体矩形を推定するタスクである~\cite{kamath2021mdetr,gupta2020contrastive}．
% マルチモーダル対話データセットは画像ではなく動画を元にした画像系列から構成される．
% すなわち，1フレーズについてグラウンディング対象の画像が複数存在する．
% 本実験では簡単のため，フレーズが含まれる発話を考え，グラウンディング対象をその発話の開始時刻から次の発話の開始時刻までの間に含まれる画像フレームに限定する．\footnote{
%   An \textit{utterance} is almost always a series of sentences uttered by a speaker before the turn transition.
%   However, in the case of a utterance with long pause, the utterance may split into multiple utterances.
% }
Given an image and a corresponding text description, phrase grounding detects the objects in the former that correspond to each phrase in the latter~\citep{kamath2021mdetr,gupta2020contrastive}.
J-CRe3 consists of image sequences from videos, not individual images.
That is, each phrase in a text (i.e., a transcribed utterance) has multiple images for grounding.
For simplicity, we considered the utterance containing the phrase and limited the grounding target to the video frames between the utterance's start and the next one's start.

% 評価指標は Recall@$k$ を使用した．
% phrase grounding model は，それぞれのフレーズについて複数の物体矩形とその予測確率を出力する．
% Recall@$k$ は正解の物体矩形のうち，出力された予測確率上位$k$件の物体矩形に含まれるものの割合である．
% ここで，先行研究~\cite{kamath2021mdetr}にならい予測された物体矩形が正解の物体矩形と 0.5 以上の Intersection-over-Union (IoU) を持つ場合に両者が一致すると判断した．
As an evaluation metric, we used Recall@$k$, which is the major evaluation metric for phrase grounding models.
It measures whether a model can rank the ``correct'' box among its top $k$ predictions.
A box is considered correct if the Intersection-over-Union (IoU) between the predicted and the ground-truth boxes exceeds a predetermined threshold.
We set the threshold to 0.5, following~\citet{kamath2021mdetr}.

% https://www.notion.so/20230612-39ccb9b9d2d34f4bbaf426880667eff8?pvs=4
% python src/evaluation.py -d data/dataset -k data/knp -i data/image_text_annotation -p result/pretrained_b3_roberta_ja_mixed_2e_mmdialogue_4e_b8 --scenario-ids $(cat test.txt) --recall-topk 1
% python src/evaluation.py -d data/dataset -k data/knp -i data/image_text_annotation -p result/pretrained_b3_roberta_ja_mixed_2e_mmdialogue_4e_b8 --scenario-ids $(cat test.txt) --recall-topk 5
% python src/evaluation.py -d data/dataset -k data/knp -i data/image_text_annotation -p result/pretrained_b3_roberta_ja_mixed_2e_mmdialogue_4e_b8 --scenario-ids $(cat test.txt) --recall-topk 10
\begin{table*}[t]
  \centering
  \small
  \begin{tabular}{l|r@{\hskip 1mm}rr@{\hskip 1mm}rr@{\hskip 1mm}r|rrr}
    \toprule
    \mc{1}{c|}{\multirow{2}{*}{\textbf{Model}}} & \mc{6}{c|}{\textbf{\ours}}   & \mc{3}{c}{\textbf{Flickr30k Entities JP}}                                                                                                                                                                                   \\% \cmidrule{2-10}
                                                & \mc{2}{c}{\textbf{Recall@1}} & \mc{2}{c}{\textbf{Recall@5}}              & \mc{2}{c|}{\textbf{Recall@10}} & \mc{1}{c}{\textbf{Recall@1}} & \mc{1}{c}{\textbf{Recall@5}} & \mc{1}{c}{\textbf{Recall@10}}                                                    \\ \midrule
    MDETR (ENB3)                                & 0.007                        & (3/407)                                   & 0.012                          & (5)                          & 0.012                        & (5)                           & 0.007          & 0.010          & 0.010          \\
    + FT1                                       & 0.214                        & (87/407)                                  & 0.381                          & (155)                        & 0.403                        & (164)                         & \textbf{0.671} & \textbf{0.819} & \textbf{0.845} \\
    + FT1 + FT2                                 & \textbf{0.337}               & (137/407)                                 & \textbf{0.474}                 & (193)                        & \textbf{0.511}               & (208)                         & 0.222            & 0.293            & 0.309           \\ \midrule
    MDETR (ENB5)                                & 0.005                        & (2/407)                                   & 0.007                          & (3)                          & 0.007                        & (3)                           & 0.018          & 0.023          & 0.024          \\
    + FT1                                       & 0.231                        & (94/407)                                  & 0.376                          & (153)                        & 0.420                        & (171)                         & \textbf{0.675} & \textbf{0.818} & \textbf{0.845} \\
    + FT1 + FT2                                 & \textbf{0.410}               & (167/407)                                 & \textbf{0.494}                 & (201)                        & \textbf{0.504}               & (205)                         & 0.215           & 0.260            & 0.265            \\
    \bottomrule
  \end{tabular}
  \caption{
    Performances of phrase grounding models: \textit{ENB3} and \textit{ENB5} indicate EfficientNet-B3 and EfficientNet-B5, which are used as backbone of MDETR.
    \textit{FT1} denotes model is fine-tuned on the existing datasets: RefCOCO, RefCOCO+, RefCOCOg, Visual Genome, GQA, and Flickr30k Entities JP.
    \textit{FT2} denotes model is further fine-tuned on \ours.
    Values in parentheses are correctly resolved references.
  }
  \label{tab:phrase-grounding-result}
\end{table*}

\subsubsection{Method}

% 学習データ不足を補うため，実験では学習済み phrase grounding モデルを既存の phrase grounding データセットを活用して fine-tuning する．
% ベースとなる phrase grounding モデルには MDETR を使用する．
% MDETR は Transformer アーキテクチャ~\cite{Transformer}に基づき，物体検出と phrase grounding を end-to-end で行うモデルである．
% しかし，訓練済み MDETRモデルには2つの問題がある．
% 1つは言語の不一致である．
% MDETR モデルは英語の phrase grounding および referring expression comprehension のデータセットで訓練されている．
% そこで我々は， Flickr30k Entities JP データセット~\cite{nakayama-tamura-ninomiya:2020:LREC}を使用してこのモデルを fine-tuning する．
% 2つ目の問題はドメインの不一致である．
% MDETR モデルは Flickr などから収集した静止画像を元に訓練されているが，本実験で使用する画像は動画のフレームである．
% したがって物体が見切れていたり，ブレていたりする．
% そこで本データセットで追加のfine-tuningをする．
To address the scarcity of training data, we fine-tuned a pre-trained phrase grounding model using existing phrase grounding datasets.
As MDETR~\citep{kamath2021mdetr}, which serves as the base phrase grounding model, performs object detection and phrase grounding in an end-to-end manner, we do not need a separate object detector.

However, a pre-trained MDETR model has two issues.
The first is language mismatch.
The MDETR model is trained on datasets for English phrase grounding and referring expression comprehension.
Therefore, we fine-tuned the model using the Flickr30k Entities JP dataset~\citep{nakayama-tamura-ninomiya:2020:LREC}, which is a Japanese translation of the Flickr30k Entities dataset~\citep{flickrentitiesijcv}, a commonly used phrase grounding dataset.

The second issue is domain mismatch.
Although the MDETR model is trained on photos intentionally taken by a human photographer, the images in our dataset are frames of egocentric video, which often contain blurred or occluded objects.
We performed additional fine-tuning on our dataset to address this issue.

\subsubsection{Training Details}
% MDETR は学習済みモデル\footnote{\url{https://github.com/ashkamath/mdetr\#pre-training}}が公開されているが，英語テキストで学習されたモデルであるため直接利用できない．
% このデータセットは phrase grounding において標準的に使用される Flickr30k Entities データセット~\cite{flickrentitiesijcv}を日本語に翻訳したものである．

% 1段階目のfine-tuningではMDETRのテキストエンコーダをRoBERTa baseからmultilingualモデルであるXLM-RoBERTaに置き換えた．
% MDETRモデル全体を新しいエンコーダに適応させるためFlickr30k Entities JPのほか，RefCOCO, RefCOCO+, RefCOCOg, Visual Genome, GQAを混合したデータセットでfine-tuningした．\footnote{In our preliminary experiment, mixing all the dataset led to better performance compared to using only Flickr30k Entities JP.}
% 学習データに含まれる言語は英語が支配的であるため日本語テキストにに対するエンコーディング能力が失われないようテキストエンコーダは最終層を除いて freeze した．
In the first stage of fine-tuning, the text encoder of MDETR, which is a RoBERTa base~\citep{RoBERTa}, was replaced with a multilingual encoder, an XLM-RoBERTa base~\citep{conneau-etal-2020-unsupervised}.
To adapt the entire MDETR model to the new encoder, we fine-tuned it with a mixture of RefCOCO~\citep{yu-2016-refcoco}, RefCOCO+~\citep{yu-2016-refcoco}, RefCOCOg~\citep{mao-2016-refcocog}, Visual Genome~\citep{krishna-2016-visualgenome}, GQA~\citep{hudson-2019-gqa}, and Flickr30k Entities JP.\footnote{In our preliminary experiment, combining all the datasets improved the performance more than using only Flickr30k Entities JP.}
As English is the dominant language in the training data, we froze the text encoder except for the final layer to prevent overfitting to English texts.

% 2段階目のfine-tuningではまず，本データセットを Flickr30k Entities と同じ形式に変換した．
% すなわち，1画像に対して10単語前後のキャプションが複数付与されている形式にした．
% 具体的には，対話テキストを2発話ずつに区切り，それぞれの2発話を1つのキャプションとして扱った．
% このときいずれの物体にも参照していない発話対は取り除いた．
% この変換後のデータセットを使用し，モデルの全パラメータを更新した．
% 学習，開発，テストにはそれぞれアノテーション済みの16，4，4対話を使用した．
In the second stage, we first converted our dataset into the same format as the Flickr30k Entities, where each image has multiple text descriptions of 10--20 words.
We divided the dialogue text into segments of two consecutive utterances and treated each one as a single text description.
Note that we removed segments without any references to objects.
Other training details are described in Appendix~\ref{sec:training-details}.

% \begin{itemize}
%   \item 訓練済みMDETRモデルには EfficientNet B3 を backbone としたモデル\footnote{\url{https://zenodo.org/record/4721981/files/pretrained_EB3_checkpoint.pth}}を使用した．
%   \item このモデルをRefCOCO, RefCOCO+, RefCOCOg, Visual Genome, GQA, Flickr30k Entities JP を混合したデータセットで fine-tuning した．Flickr30K Entities JP はFlickr30k Entitiesデータセットを日本語に翻訳したもので，31,000枚の画像とそれぞれに5つ付随するキャプションが含まれる．
%   \item text encoder は XLM-RoBERTa\footnote{\url{https://huggingface.co/xlm-roberta-base}}を使用した．
%   \item 学習データに含まれる言語は英語が支配的であるため日本語に対する性能を落とさないよう text encoder は最終層を除いて freeze した．
%   \item 訓練は2エポック行った．Flickr30kEnt-JPのスコアを比較すると1エポック目よりも2エポック目の方がわずかに良かった．
%   \item 本データセットでのfine-tuningの際は text encoder も含めて全てのパラメータを更新した．
%   \item fine-tuningのため，本データセットを Flickr30k Entities と同じ形式に変換した．すなわち，1画像に対して10単語前後のキャプションが複数付与されている形式にした．具体的には，対話テキストを2発話ずつに区切り，それぞれの2発話を1つのキャプションとして扱った．このときいずれの物体にも参照していない発話対は取り除いた．
%   \item 学習にはタグ付与済みの53対話のうち38対話を使用した．残りの対話のうち6対話は評価データに，9対話はテストデータに使用した．
% \end{itemize}

\begin{figure}[t]
  \centering
  \includegraphics[width=0.48\textwidth]{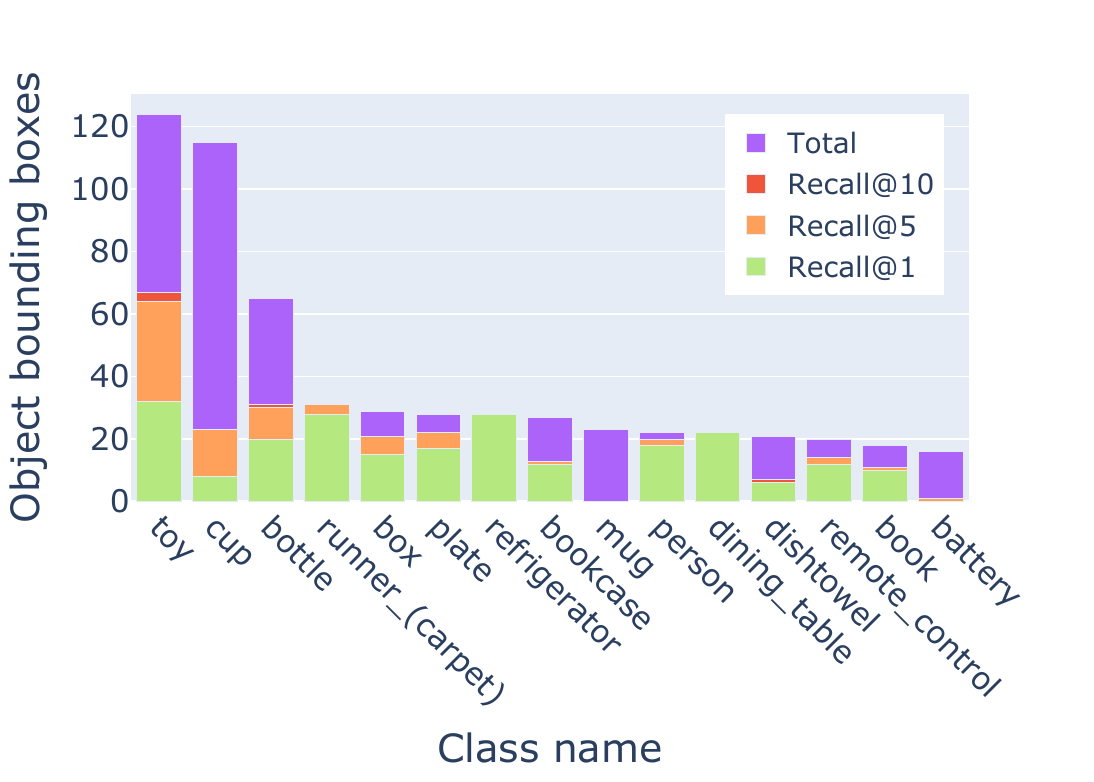}
  \caption{
    Distribution of Recall@$k$ and number of objects for each object class: Figure shows top 15 classes.
    Although we show the difference between Recall@5 and Recall@10 in red in the figure, there were very few of them.
  }
  \label{fig:recall-distribution}
\end{figure}

\begin{figure}[t]
  \centering
  \includegraphics[width=0.45\textwidth]{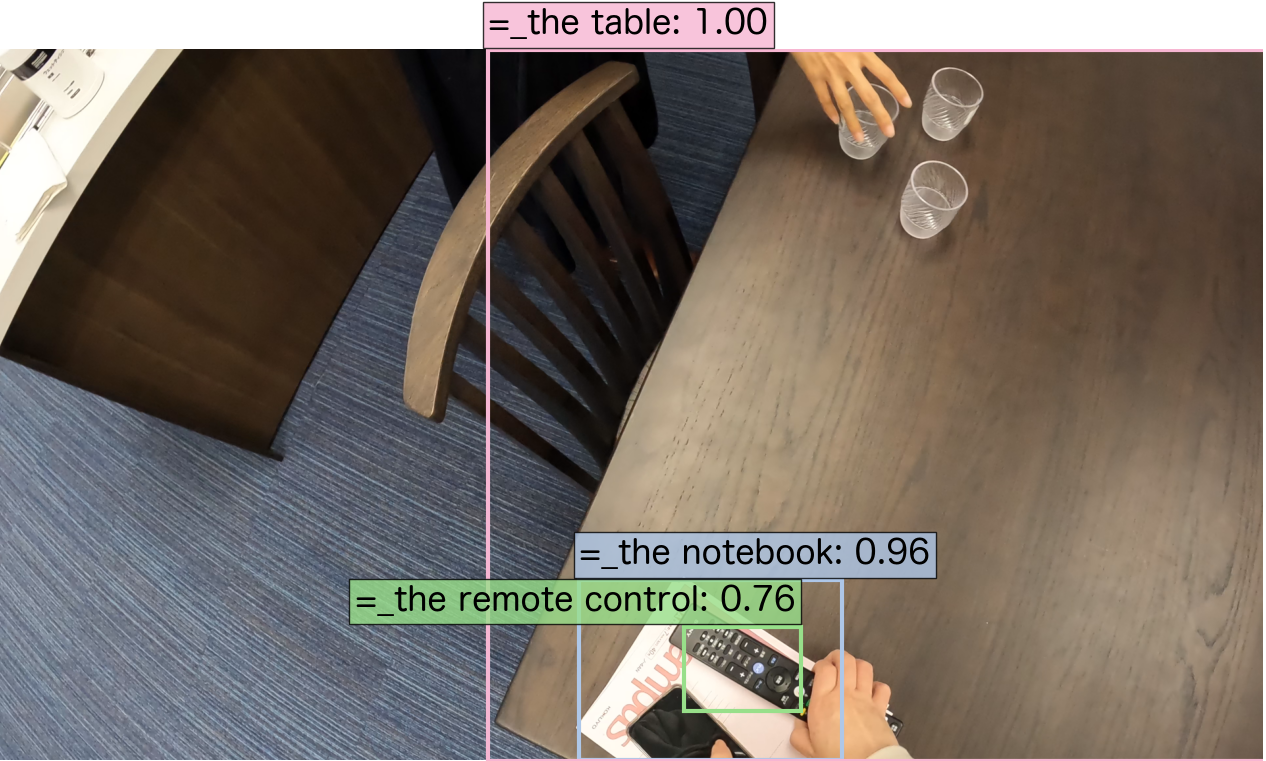}
  \caption{
    % 「とーノートと携帯は隣の部屋の机の上リモコンはソファの上ですよね。」という発話に対する phrase grounding モデルの解析例．
    % 物体矩形にはフレーズとシステムの予測確率が記載されている．
    Example of phrase grounding for utterance ``The notebook and mobile phone should be put on the table in the next room, and the remote control should be put on the sofa, right?'' (translated).
    Predicted phrases and their confidences are shown with colored object bounding boxes.
    The confidence threshold is 0.5.
  }
  \label{fig:case-analysis}
\end{figure}

\subsubsection{Results}
% 実験結果を表~\ref{tab:phrase-grounding-result}に示す．
% fine-tuningはいずれも本データセットにおける性能向上に寄与している．
% しかし，Flickr30k Entities JP と比較すると Recall@$k$ のスコアは大幅に低い．
% このことから，本データセットは既存のデータセットと比べドメインが大きく異なり，既存手法では解くことが難しいことが分かる．
Table~\ref{tab:phrase-grounding-result} shows the Recall@$k$ of the phrase grounding model.
The first-stage fine-tuning contributes to both the \ours~and Flickr30k Entities JP performances, suggesting that it largely mitigates the language mismatch issue.
The second-stage fine-tuning also improved the \ours~performance.
However, compared to Flickr30k Entities JP, the scores are significantly lower, which suggests that domain mismatch remains a significant issue.

% 図~\ref{fig:recall-distribution}に本データセットのクラスごとの Recall の分布を示す．
% Recall が低いクラスとして cup, mug, book, battery などが挙げられる．
% これら物体は比較的小さく，また，同じ物体が複数出現することが多いため性能が低いと考えられる．
Figure~\ref{fig:recall-distribution} visualizes the distribution of Recall@$k$ for each object class.
The classes with low recall include \textit{cup}, \textit{mug}, and \textit{battery}.
One possible reason is that objects in these classes are relatively small and often appear with similar objects in the same class.
Generally, detecting small objects is known to be a challenging task~\citep{rekavandi2023transformers}.
Using dedicated object detectors instead of MDETR possibly improves the coverage of detected objects and the performance of phrase grounding.

% 図~\ref{fig:case-analysis}に我々のデータセットにおけるモデルの解析例を示す．
% 「ノート」「リモコン」「テーブル」という独立した物体名に注目すると，それぞれの物体を正しく特定できていることが分かる．
% しかし，「テーブル」は「隣の部屋のテーブル」と言及されており，文脈を考慮すると正しくないことが分かる．
% 高度な文脈理解を必要とすることは本データセットにおける phrase grounding の難しさ要因の一つである．
Figure~\ref{fig:case-analysis} shows an example of the phrase grounding results on our dataset.
When we focus on such object names as \textit{the notebook}, \textit{the remote control}, and \textit{the table}, the bounding boxes seem correctly identified.
However, \textit{the table} is referred to as \textit{the table in the next room}, which is not shown in the frame.
Although we trained the model not to ground phrases whose corresponding objects did not appear in the frame, the model grounded \textit{the table} with high confidence.
Another problem is the failure to ground the phrase \textit{mobile phone} to the smartphone beside the remote control.
Indeed, identifying a smartphone from this image alone is challenging due to insufficient visual information.
Requiring a high level of textual and visual context understanding is one of the factors that make phrase grounding on our dataset challenging.

\subsubsection{Impact of the Behavior of the Robot Actor on the Model Performance}
% 本データセットはロボットと人間の対話を想定するが，収録においてはロボット役を人間の演者に依頼している．
% そのため，視線移動や物体把持などの行動に人間の思考が介在し，解析モデルにとって都合の良い1人称視点動画が得られている可能性がある．
% 例えば，複数のコップが並んでいる場面で，「一番大きなコップに水を入れて」と主人が発話した場合，ロボット役演者は一番大きなコップに目を向けそれを手に取ることが予想される．
% 演者がコップを手に取った後の動画フレームには，そのコップが大きく映っていることが期待され，解析モデルにとって「一番大きなコップ」の参照先を特定することは容易となる．
The conversations in our dataset are intended to be between a robot and a human; however, in its recording, human actors are employed to play the roles of robots.
Thus, human thoughts interfere in the actor's behaviors, such as gaze shift and object grasping, potentially resulting in egocentric videos that are conveniently analyzable for the model.
For instance, in a scene with multiple cups, if the master says, ``Fill the biggest cup with water,'' the robot actor will look at and pick up the biggest cup.
After the actor picks up the cup, the video frame is expected to prominently feature that cup, making it easier for the model to identify the ``biggest cup'' as the referent.

\begin{table}[t]
  \centering
  \small
  \scalebox{0.81}{
    \begin{tabular}{l|r@{\hskip 1mm}rr@{\hskip 1mm}rr@{\hskip 1mm}r}
      \toprule
      \textbf{Temporal loc.} & \mc{2}{c}{\textbf{Recall@1}} & \mc{2}{c}{\textbf{Recall@5}} & \mc{2}{c}{\textbf{Recall@10}}                                  \\
      \midrule
      First half             & 0.361                        & (110/305)                    & 0.472                         & (144) & 0.489          & (149) \\
      Second half            & \textbf{0.366}               & (120/328)                    & \textbf{0.500}                & (164) & \textbf{0.512} & (168) \\
      After                  & {0.313}                      & (60/192)                     & 0.396                         & (76)  & 0.417          & (80)  \\
      \bottomrule
    \end{tabular}
  }
  \caption{
    % 動画フレームの時間的位置とRecallの関係．
    % テストセットと開発セットを合わせた18対話で評価した．
    Relation between the temporal location of the video frames and Recall@$k$.
    We evaluated 18 dialogues, including the validation and test sets.
  }
  \label{tab:temporal-location}
\end{table}

\begin{table*}[t]
  \centering
  \small
  \begin{tabular}{l|r@{\hskip 1mm}rr@{\hskip 1mm}rr@{\hskip 1mm}r}
    \toprule
    \mc{1}{c|}{\textbf{Task}}         & \mc{2}{c}{\textbf{Recall@1}} & \mc{2}{c}{\textbf{Recall@5}} & \mc{2}{c}{\textbf{Recall@10}}                         \\ \midrule
    Nominative reference resolution   & 0.064                        & (79/1230)                    & 0.070                         & (86)  & 0.072 & (88)  \\
    Accusative reference resolution   & 0.199                        & (67/336)                     & 0.232                         & (78)  & 0.235 & (79)  \\
    Dative reference resolution       & 0.035                        & (25/719)                     & 0.047                         & (34)  & 0.047 & (34)  \\
    Nominative-2 reference resolution & 0.000                        & (0/399)                      & 0.000                         & (0)   & 0.000 & (0)   \\
    Bridging reference resolution     & 0.198                        & (17/86)                      & 0.198                         & (17)  & 0.198 & (17)  \\
    Direct reference resolution       & 0.410                        & (167/407)                    & 0.494                         & (201) & 0.504 & (205) \\
    \bottomrule
  \end{tabular}
  \caption{
    Result of text-to-object reference resolution:
    Numbers in parentheses denote correctly resolved references and their total number.
  }
  \label{tab:visual-reference-result}
\end{table*}

% 演者の行動の解析精度への影響を調査するため，発話に対する解析対象フレームの時間的な位置ごとにモデルの性能を評価した．
% 演者の行動は相手の発話を受けて行われるため，モデルが演者の行動を手がかりにしている場合，時間的に後ろのフレームに対する性能が高くなると考えられる．
% フレームをその時間的な位置に基づき以下の3種類に分類し，それぞれのカテゴリにおけるRecall@$k$を評価した．
To investigate the impact of the actor's behavior on the model performance, we evaluated the phrase grounding model with respect to the temporal location of the frames corresponding to the target utterance.
If the model uses the actor's behaviors as clues, its performance is better on later frames as the actor's actions are responses to the master's utterances.
We classified the frames into three categories based on their temporal locations and evaluated Recall@$k$ for each category.

\begin{itemize}
  % \item 発話開始時点から発話区間中央までの間のフレーム（発話中前半）
  % \item 発話区間中央から発話終了時点までの間のフレーム（発話中後半）
  % \item 発話終了時点から次の発話開始時点までの間のフレーム（発話以降）
  % \item Frames between the end of the previous utterance and the start of the target utterance
  \item Frames temporally corresponding the first half of the target utterance (\textbf{First half})
  \item Frames temporally corresponding the second half of the target utterance (\textbf{Second half})
  \item Frames between the end of the target utterance and the start of the next utterance (\textbf{After})
\end{itemize}

% 表~\ref{tab:temporal-location}に結果を示す．
% 発話中前半に比べ，発話中後半や発話以降のフレームに対する性能が高く，演者の行動が解析モデルにとっての手がかりになっていることが示唆された．
% したがって，本データセットを用いてシステムを評価する際には，発話中前半のフレームにおける性能を重視する必要がある．
% また，実世界で他者と協働するロボットにおいては，視線移動等の行動を正しく行う能力が重要であることも示唆される．
Table~\ref{tab:temporal-location} shows the results.
The performance is higher for the Second half frames than the First half ones, suggesting that the actor's behavior serves as a clue for the model.
Therefore, when evaluating a system using our dataset, it is important to focus on the performance for the earlier frames corresponding to the target utterance.
% This result also suggests that the ability to perform actions such as gaze shift is important for robots that collaborate with humans in the real world. そもそも正しく gaze shift ができるならグラウンディングはすでに解けているのでは？

\subsection{Combining the Results}

% phrase grounding モデルは間接的な参照関係を扱えない．
% すなわち，述語に対してそのヲ格に相当する物体を特定したり，名詞に対して橋渡し照応関係を持つ物体を特定することができない．
% しかし，phrase grounding モデルの出力結果とテキスト間照応解析モデルの出力結果を統合することで，これら解析が可能になる．
The phrase grounding model cannot handle indirect reference relations.
In other words, it cannot identify an object that corresponds to a predicate's arguments, nor can it identify an object that has a bridging reference relation with a noun.
However, we can resolve these references by combining the output of the phrase grounding model with that of the textual reference resolution model.

% 表~\ref{tab:visual-reference-result}に2つのモデルを組み合わせた場合の物体検出およびテキスト・物体間参照解析の結果を示す．
% エラー伝播の問題から，いずれのタスクにおいても性能は著しく低い．
% 特に，phrase grounding の Recall@1 が0.318 でありこの値が上界になっている影響が大きい．
Table~\ref{tab:visual-reference-result} shows the combined results of the object detection and text-to-object reference resolution.
The performance on all tasks is significantly low due to error propagation issues.
In particular, the Recall@1 of the phrase grounding model is 0.410, which is the upper bound of the performance and greatly impacts the results.

\section{Related Work}

% 人間と物体とのインタラクションを含む一人称動画データセットとしてはEgo4D（引用）やEPIC-Kitchens（引用）やHome Action Genome（引用）やBioVL2（引用）が挙げられる．
% Ego4D, EPIC-Kitchens, Home Action Genome は我々のデータセットと同じく日常動作を含む一方，BioVL2は生化学分野の実験動画を含む．
% これらデータセットは動作認識などの動画に対する粗い特徴が付与されており，参照表現の接地は扱っていない．
% BioVL2は物体矩形が付与されているが，対象となる物体は実験プロトコルに存在し，かつ手と触れているものに限られる．
% % さらに BioVL2は対話ではなく実験プロトコルのテキストと紐づいており，ゼロ照応も扱わない．
Egocentric video datasets that involve human-object manipulation tasks include Ego4D~\citep{Ego4D2022CVPR}, EPIC-Kitchens~\citep{Damen2022RESCALING}, Home Action Genome~\citep{Rai-2021-HomeActionGenome}, and BioVL2~\citep{nishimura-2021-iccvw,nishimura-2022-jnlp}.
Ego4D, EPIC-Kitchens, and Home Action Genome feature everyday actions similar to our datasets; BioVL2 contains experimental videos from the biochemistry field.
These datasets focus on actions in videos and do not address the localization of objects based on dialogue.
Although BioVL2 provides bounding boxes, the target objects are restricted to those in the experimental protocol and in contact with the experimenter's hand.

% テキストと画像中の物体矩形を結びつけるタスクとしては， や referring expression comprehension (REC) やphrase grounding が挙げられる．
% REC は与えられた text description に対応する物体領域を検出するタスクである．
% phrase groundingはさらに挑戦的なタスクであり， text description に含まれるすべてのフレーズについて対応する物体領域を検出する．
% しかし，いずれのタスクもテキストと直接的な関係を持つ物体領域しか扱わない．
% 一方，マルチモーダル照応解析は，橋渡し照応関係や述語項構造などの間接的な関係まで包含する．
The tasks that associate text-to-object bounding boxes include referring expression comprehension (REC, ~\citealp{yu-2016-refcoco}) and phrase grounding~\citep{flickrentitiesijcv}.
REC detects object regions that corresponds to a given text description.
Phrase grounding is more challenging, since it involves detecting object regions for all the phrases within a text description.
However, both tasks only address object regions directly related to the text.
In contrast, multimodal reference resolution encompasses indirect relations, such as bridging reference relations and predicate-argument structures.

\section{Conclusion}

We proposed a multimodal reference resolution task to realize a robot that interacts and collaborates with humans in the real world.
This task consists of textual reference resolution, which identifies the phrases that are referred to in texts; object detection, which detects referent object candidates in images; and text-to-object reference resolution, which identifies referent objects from the output of object detection.
To train and evaluate models for this task, we constructed a Japanese Conversation dataset for Real-world Reference Resolution (\ours).
It is based on real-world conversations and is expected to contribute to the development of more practical dialogue understanding systems.

Our future plans include improving the resolution model for this task.
Although the system used in our experiments analyzed textual and text-to-object references independently, resolving these relations in an integrated manner could improve the performance.
% In addition, as this dataset contains videos, it is also possible to use time series information over frames.

Another way to improve the resolution model is to expand the size and domain of the dataset.
We can use text or image generation models to expand the dataset at a low cost.
Specifically, it is possible to append generated dialogues to images in existing phrase grounding datasets.
% To improve the textual reference resolution model, especially for the accusative case, we plan to take visual information into account.

\section*{Limitations}

Constructing a real-world conversation dataset is costly, and our dataset is limited in size, making it insufficient for independently training a model.
However, this issue can be mitigated by combining it with other datasets or pre-trained models.

Our dataset is designed for interactions between a master and a domestic assistant robot.
Consequently, the model's generalizability to other contexts remains uncertain, such as outdoor environments, or different types of human-robot interactions, such as interactions with multiple robots.

\section*{Acknowledgements}
% 本研究は，京都大学科学技術イノベーション創出フェローシップ事業の助成を受けたものである．
% 本研究の一部はJSPS科研費22H03654の支援を受けたものである．
This work was supported by Kyoto University Science and Technology Innovation Creation Fellowship (Information / AI field).
This work was partially supported by JSPS KAKENHI Grant Number 22H03654.

\bibliography{main}
\bibliographystyle{lrec-coling2024-natbib}

% \section{Language Resource References}\label{lr:ref}
% \bibliographystylelanguageresource{lrec-coling2024-natbib}
% \bibliographylanguageresource{languageresource}

\appendix

\onecolumn
% {付録のサンプル:付録は独立に1ページだけ}
% %\small % 文字サイズを小さくする． 一行23文字 => 26文字
\section{Scenario Example}\label{sec:scenario-example}
% 表~\ref{tab:scenario}に収集した対話シナリオの例を示す．
% 括弧内のテキストは場面状況を表す．
% 場面状況は対話収録にのみ使用し，データセットには含まれない．
% 括弧外の発話には「あそこ」や「それ」などの参照表現が含まれ，視覚情報も含めなければ理解が困難な対話になっている．
Table~\ref{tab:scenario} shows an example of our collected scenarios.
The text in the parentheses describes the scene contexts, which were used only for conversation recording and were not included in our dataset.
Texts outside the parentheses contain such referential expressions as \textit{over there} and \textit{it}, making it difficult to understand the dialogue text without visual contexts.

\begin{table*}[h]
  \centering
  \scalebox{0.8}{
    \begin{tabular}{ll}
      \toprule
      \textbf{Speaker} & \textbf{Utterance}                                                                                                    \\
      \midrule
      % Master  & \Ja{人形を梱包したいんだけど、紙をシュレッダーにかけて緩衝材を作ってくれない？}           \\
      Master  & I want to pack this doll. Can you shred some paper and make some cushioning material for it?                 \\
      % Robot   & \Ja{分かりました。どの紙を使ったらいいですか？}                           \\
      Robot   & Yes, I can. What paper should I use?                                                                         \\
      % Master  & \Ja{（部屋の隅の雑誌の山を指差して）あそこから適当に使って。その段ボール箱に一杯分くらい。}     \\
      Master  & Use that paper (points at a pile of magazines in the room's corner). Take enough to fill that cardboard box. \\
      % Robot   & \Ja{（古雑誌を黙々とシュレッダーにかける）このくらいでよろしいですか？}               \\
      Robot   & (Shreds old magazines) Is this enough?                                                                       \\
      % Master  & \Ja{（段ボールの中を確認しながら）うん、大丈夫。そしたら、そこにあるプチプチを持ってきてもらえる？} \\
      Master  & (Checks inside the cardboard box) Yeah, that's plenty. Can you bring me that bubble wrap over there?         \\
      % Robot   & \Ja{分かりました。（プチプチを渡す）ついでに、それも包みましょうか？}                \\
      Robot   & Yes, of course (Hands it over) Should I wrap it up as well?                                                  \\
      % Master  & \Ja{ううん、壊れ物だから自分で包むよ。（人形を包む）テープを持ってきて。}              \\
      Master  & No, it's too fragile, let me do it. (Wraps the doll) Bring me some tape.                                     \\
      % Robot   & \Ja{テープはどこにありますか？}                                   \\
      Robot   & Where is it?                                                                                                 \\
      % Master  & \Ja{（棚を指差して）たぶん右の戸棚に入っていると思う。（人形を箱に入れる）}             \\
      Master  & (Pointing to a shelf) I think it's probably in the cupboard on the right. (Places the doll in the box)       \\
      % Robot   & \Ja{（棚に向かい、クラフトテープを掴み）これでよろしいですか？}                   \\
      Robot   & (Goes to the shelf and grabs some wrapping tape) Is this okay?                                               \\
      % Master  & \Ja{うん。人形は詰めたから、あとはテープで蓋を閉じて、玄関先まで運んでおいて。}           \\
      Master  & Yes. I've packed the dolls, so tape the lid and carried it to the front door.                                \\
      % Robot   & \Ja{分かりました。（段ボールをテープで閉じる）作業が終わりましたので、箱を玄関先に置いてきます。}  \\
      Robot   & I understand (The box is taped shut). I'm done, so I'll leave the box at the front door.                     \\
      \bottomrule
    \end{tabular}
  }
  \caption{
    Example of collected scenarios:
    Text in parentheses is scene context and was not used as dialogue.
  }
  \label{tab:scenario}
\end{table*}

\section{Training Details}\label{sec:training-details}

We trained the phrase grounding model with the hyper-parameters shown in Table~\ref{tab:hyper-parameters-1}, following~\citet{kamath2021mdetr}.
The EfficientNet-B3 and EfficientNet-B5 models were downloaded from \url{https://github.com/ashkamath/mdetr?tab=readme-ov-file#pre-training}.
The XLM-RoBERTa base model was downloaded from \url{https://huggingface.co/xlm-roberta-base}.

\begin{table}[h]
  \centering
  \small
  \begin{tabularx}{0.65\textwidth}{l| *{2}{Y}}
    \toprule
    Settings                           & FT1                                                                                          & FT2 \\
    \midrule
    Detection Backbone                 & \multicolumn{2}{c}{EfficientNet-B3 or EfficientNet-B5}                                             \\
    Text Encoder                       & \multicolumn{2}{c}{XLM-RoBERTa base}       \\
    Batch size                         & \multicolumn{2}{c}{8}                                                                              \\
    Training Epochs                    & \multicolumn{2}{c}{2}                                                                              \\
    Learning Rate                      & \multicolumn{2}{c}{1e-4}                                                                           \\
    Learning Rate (detection backbone) & \multicolumn{2}{c}{5e-5}                                                                           \\
    Learning Rate (text encoder)       & \multicolumn{2}{c}{5e-5}                                                                           \\
    Weight Decay                       & \multicolumn{2}{c}{1e-4}                                                                           \\
    Gradient Clipping                  & \multicolumn{2}{c}{0.1}                                                                            \\
    Exponential Moving Average Decay   & \multicolumn{2}{c}{0.9998}                                                                         \\
    \bottomrule
  \end{tabularx}
  \caption{Hyper-parameters used for fine-tuning phrase grounding model.}
  \label{tab:hyper-parameters-1}
\end{table}

\section{Crowdsourcing Interface}\label{sec:crowdsourcing-interface}

\begin{figure}[t]
  \centering
  \includegraphics[width=0.75\textwidth]{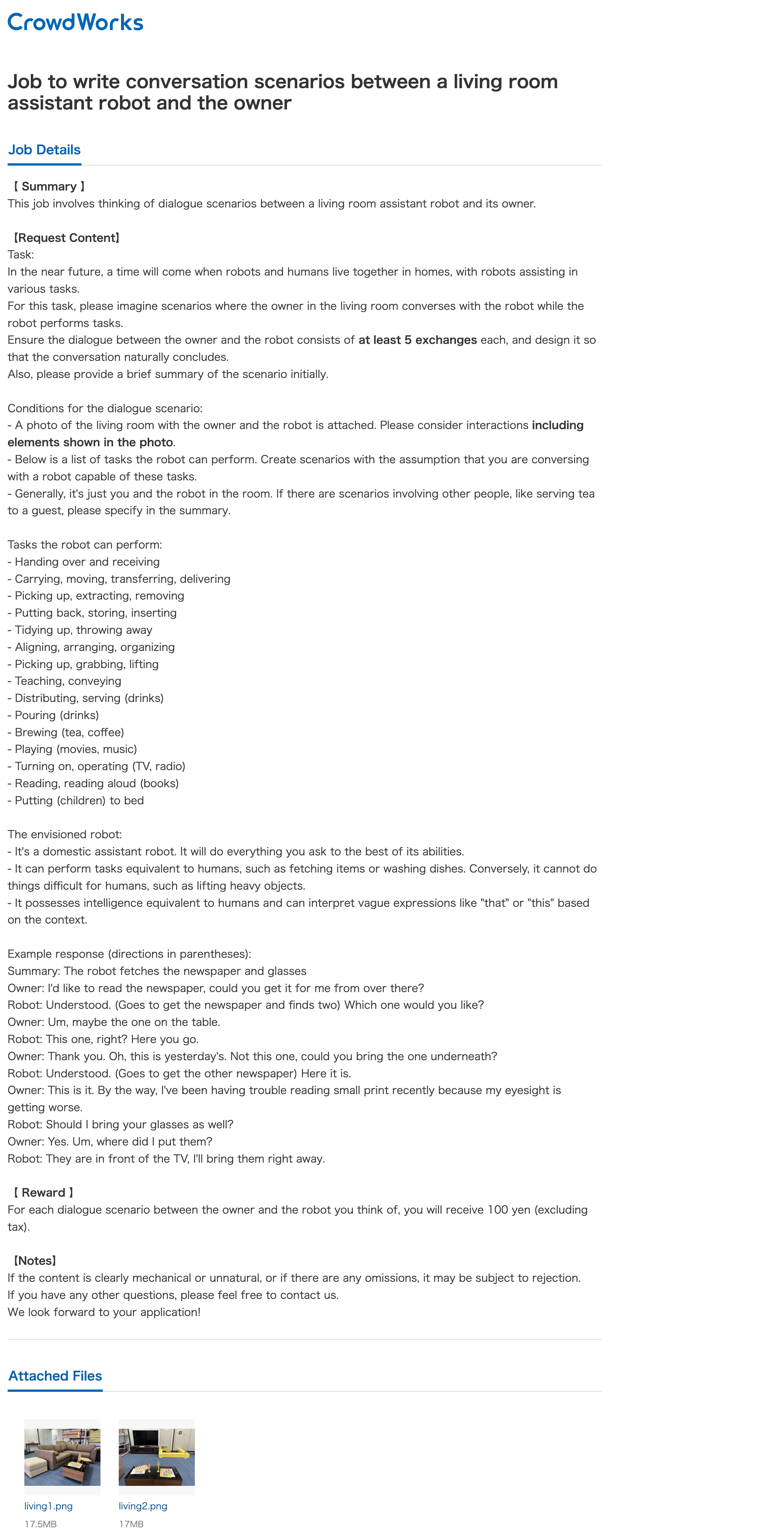}
  \caption{
    Translated crowdsourcing interface used for collection of scenarios in living room
  }
  \label{fig:crowdsourcing-interface}
\end{figure}

Figure~\ref{fig:crowdsourcing-interface} shows the crowdsourcing interface used for the scenario collection.
Note that the interface shown is an English translation of the original Japanese interface.

\end{document}